\documentclass[runningheads]{llncs}
\usepackage{graphicx}

\usepackage[table]{xcolor}
\usepackage{tabularx}
\usepackage{amsmath,amssymb,stmaryrd}  
\usepackage{array,multirow}
\usepackage{breqn}
\usepackage{graphicx}
\usepackage{tikz}
\usepackage{booktabs}
\usepackage{pifont}

\newcommand{\xmark}{\ding{53}}%

\begin{document}

\pagestyle{headings}
\mainmatter

\title{Pyramidal Edge-maps and Attention based Guided Thermal Super-resolution} 


\titlerunning{PAGSR: Pyramidal edge-maps and Attention based GSR}
\author{Honey Gupta\orcidID{0000-0002-2803-622X} \and
Kaushik Mitra\orcidID{0000-0001-6747-9050}}
\authorrunning{H. Gupta and K. Mitra}
\institute{Computational Imaging Lab, IIT Madras, India\\
\email{hn.gpt1@gmail.com; kmitra@ee.iitm.ac.in}\\
}

\maketitle

\begin{abstract}
Guided super-resolution (GSR) of thermal images using visible range images is challenging because of the difference in the spectral-range between the images. This in turn means that there is significant texture-mismatch between the images, which manifests as blur and ghosting artifacts in the super-resolved thermal image. To tackle this, we propose a novel algorithm for GSR based on pyramidal edge-maps extracted from the visible image. Our proposed network has two sub-networks. The first sub-network super-resolves the low-resolution thermal image while the second obtains edge-maps from the visible image at a growing perceptual scale and integrates them into the super-resolution sub-network with the help of attention-based fusion. Extraction and integration of multi-level edges allows the super-resolution network to process texture-to-object level information progressively, enabling more straightforward identification of overlapping edges between the input images. Extensive experiments show that our model outperforms the state-of-the-art GSR methods, both quantitatively and qualitatively.

\end{abstract}

\section{Introduction}
\label{sec:intro}

Thermal imaging has many advantages over traditional visible-range imaging as it works well in extreme visibilty conditions. It has found applications in various fields such as firefighting \cite{fire}, gas leakage detection \cite{gas}, and automation \cite{odo1,odo2,odo3}, but the high cost of thermal sensors has considerably restricted its consumer application. Super-Resolution~(SR) techniques can increase its applicability by simulating accurate high-resolution thermal images from measurements captured from the considerably inexpensive low-resolution thermal cameras. 

Efficient methods have been proposed to perform super-resolution directly from the low-resolution thermal measurements. These single image SR methods \cite{t1,t2,msf,ten,zhang2018infrared} either take an iterative approach~\cite{zhang2018infrared}
or use a convolutional neural network (CNN) to learn the upsampling transformation function $\Psi$, such that $x_h = \Psi(x_l)$.
However, if the dimensions of the input thermal image are very small, for \textit{e.g.} the thermal images from a low-end thermal camera FLIR-AX8 have a resolution of $60\times80$, then single image super-resolution becomes very challenging as the problem becomes highly ill-posed.

\begin{figure}
    \includegraphics[width=\linewidth]{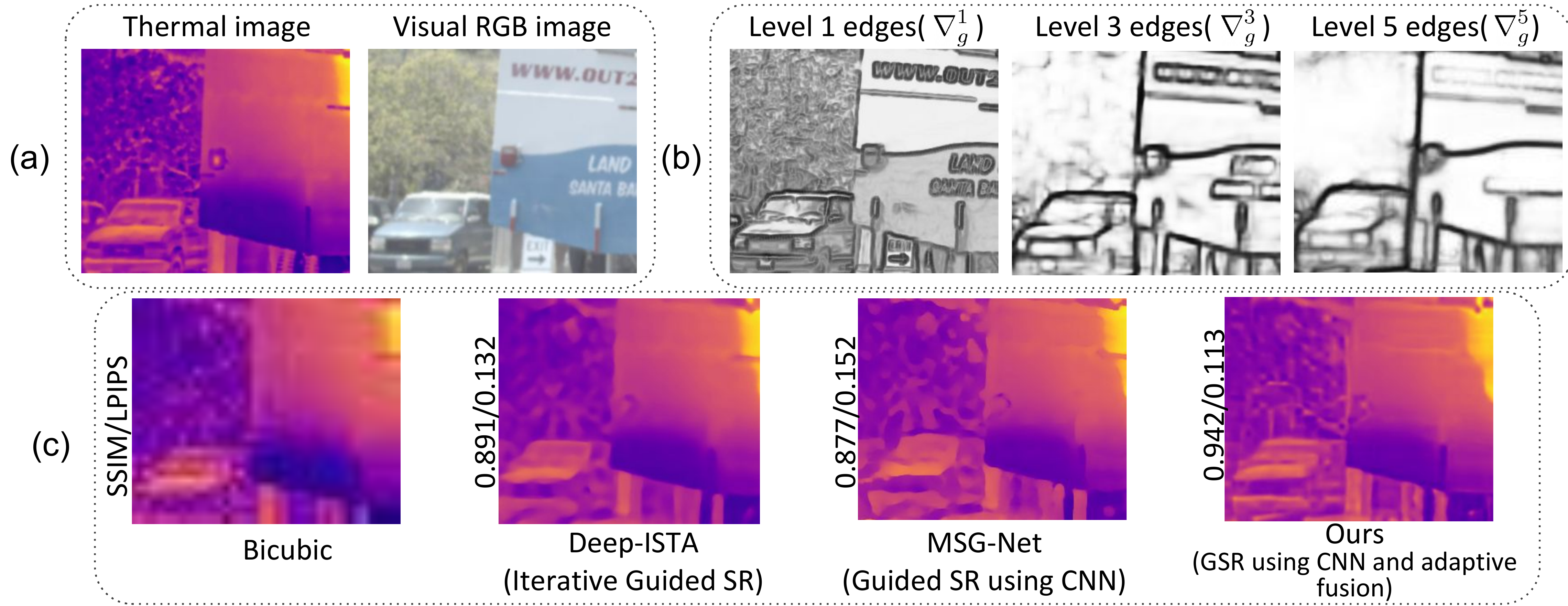}
    \caption{(a) Texture difference between thermal and visible images. (b) Multi-level edge-maps extracted from the visible image. Finer high-frequency details are present in Level-1 edges, but this level also contains the unwanted edges such as \textit{texts on the back of the truck}, which are absent in Level-5.  
    This antithetical variation of high-frequency information motivates the use of multi-level edge-maps.
    (c) Due to texture-mismatch, some existing methods such as MSG-Net~\cite{msgnet} can produce blurred images. However, with the help of pyramidal edge-maps and adaptive fusion, our method is able to produce better high-frequency details. }
    \label{fig:intro}
\end{figure}

Since many low-resolution thermal cameras are accompanied by a high-resolution visible-range camera, a practical solution to get better super-resolved thermal images is to use Guided Super-Resolution~(GSR) techniques. A crucial part of super-resolution is to correctly predict the high-frequency details. These high-frequency details are present as edge information in the two images. Since the edges are shared across the modality, estimation of the overlapping edges between the two modalities becomes a crucial task for reconstructing better high-frequency details. Most of the existing guided thermal SR techniques~\cite{brightness,ir,multimodal,chen2016color,ni} estimate high-frequency details implicitly by using CNNs and end-to-end learning.
However, the RGB guide image contains fine-texture details that are confined to the visible spectrum. For e.g., the \textit{texts on the back of the truck} in Fig.\ref{fig:intro}(a) are present only in the visible-range image. 
Such non-overlapping texture details can cause artifacts when used for guided super-resolution of thermal images. To address this drawback of single RGB guide images, we propose to use hierarchical edge-maps as guide input to our thermal super-resolution network.

We propose a GSR model that takes multi-level edge-maps extracted from the visible image as input instead of a single RGB guide image. Edge-maps at different perceptual scale contain fine-texture to object-level information distinctly, as shown in Fig.~\ref{fig:intro}(b). Due to this property of multi-level edge-maps, they can enhance the GSR performance as edge-maps at different scales could allow a more straightforward estimation of overlapping edge information. Furthermore, to allow the network to adaptively select the appropriate edge information for the input multiple edge-maps, we propose a spatial-attention based fusion module. This module adaptively selects the high-frequency information from the edge-maps before integrating them into the SR network at different depths or receptive-field sizes. Extensive experiments show that using such hierarchical form of guidance input followed by adaptive attention-based integration helps reconstruct better high-frequency details and enhances the SR performance. Our experiments also indicate that using hierarchical edge-maps and adaptive fusion can provide some robustness towards small geometric misalignment between the input images. 
In summary, the main contributions of this paper are:


\begin{itemize}
    \item We propose a novel guided super-resolution method that consists of two sub-networks: one for thermal image super-resolution and the other for feature extraction and integration of multi-level edge-maps obtained from the visible image.
    
    \item We use hierarchical edge-maps as guidance input and propose a novel fusion module that adaptively selects information from these multi-level edge-maps and integrates them into our SR network with the help of spatial-attention modules.
    
    \item We compare our model with existing state-of-the-art GSR methods and show that our method reconstructs more high-frequency details and performs significantly better, both quantitatively and perceptually.
\end{itemize}


\section{Related Works}
The high cost of thermal cameras has in the past inspired many research works to aim at thermal image super-resolution. Among the single thermal image super-resolution methods \cite{ten,t1,t2,msf}, Choi \textit{et al.} \cite{ten} suggested a shallow three-layer convolutional neural network (CNN). Zhang \textit{et al.} \cite{zhang2018infrared} thereafter combined compressive sensing and deep learning techniques to perform infrared super-resolution. Apart from thermal super-resolution methods, multiple methods have been proposed for near-infrared image super-resolution \cite{ir,ir2,ir3,g1,multimodal}. 
However, the drawback of these methods is that they do not target super-resolution for very low-resolution inputs, which is the case for low-cost thermal cameras. Choi \textit{et al.} \cite{ten} and Lee \textit{et al.}'s \cite{brightness} works suggest that using a visible super-resolution method or pre-trained model should perform well in the case of thermal images too. 
Many deep CNN based single image super-resolution methods, such as \cite{sr1,sr2,sr3,sr4,sr5,sr6,sr7,sr8,enet,sr10,rdn,rcan,san}, have shown great performance on visible images. Most of the recent methods such as RCAN\cite{rcan} and SAN \cite{san} use self-attention mechanism to produce better reconstructions. But the common concern related to single image methods is that reconstructing HR images solely from low-resolution noisy sensor images can be challenging and a guided approach might perform better.\\

Among the guided thermal super-resolution methods, Lee \textit{et al.}~\cite{brightness} used brightness information for the visible-range images. Han \textit{et al.}~\cite{ir} proposed a guided super-resolution method using CNN that extracts features from infrared and visible images and combines them using convolutional layers. Ni \textit{et al.}~\cite{ni} proposed a method to utilize an edge-map and perform GSR. Almasri \textit{et al.}\cite{msf} performed a detailed study of different CNN architectures and up-sampling methods and proposed a  network for guided thermal super-resolution. Interestingly, many recent methods for  guided super-resolution for depth \cite{depth16,depth17,depth19,depth,joint,aniso,d5,d6,d7,d8,d9} or hyperspectral \cite{hs1,hs2,hs3,hs4} images have similar backbone as the thermal guided super-resolution methods. They all use some variation of the Siamese network~\cite{siamese} to simultaneously extract information from both images and merge them to reconstruct the super-resolved image. However, these methods tackle texture-mismatch with the help of implicit or end-to-end learning, which can perform sub-optimally and lead to blurred reconstructions, as shown in Fig.~\ref{fig:intro}(c). 



\section{Pyramidal Edge-maps and Attention based Guided Super-Resolution (PAG-SR)}

The guide image belongs to a higher resolution as compared to the input thermal image and has useful high-frequency details that can be fused with the low-resolution thermal image to perform better super-resolution. However, these high-frequency details should be extracted and integrated adaptively according to the input low-resolution thermal image. Non-optimality in feature-extraction can propagate the texture-mismatch, which can further cause artifacts in the reconstructed image. At a first glance, it seems that extracting the object-level edges could be an ideal solution as they are shared across the multispectral images, but as one can observe from Fig.~\ref{fig:intro}(b), there are high-frequency details present in edge-maps at lower levels, \textit{i.e.} levels 1 and 3 that are equally useful. To resolve this conundrum, we use edge-maps extracted at pyramidal levels from the visible image and integrate them with the help of adaptive fusion module using self-attention mechanism. This way, the network can leverage high-frequency information in a hierarchical fashion and adaptively select appropriate features according to the input low-resolution thermal image.

\begin{figure}
\begin{center}
\includegraphics[width=\linewidth]{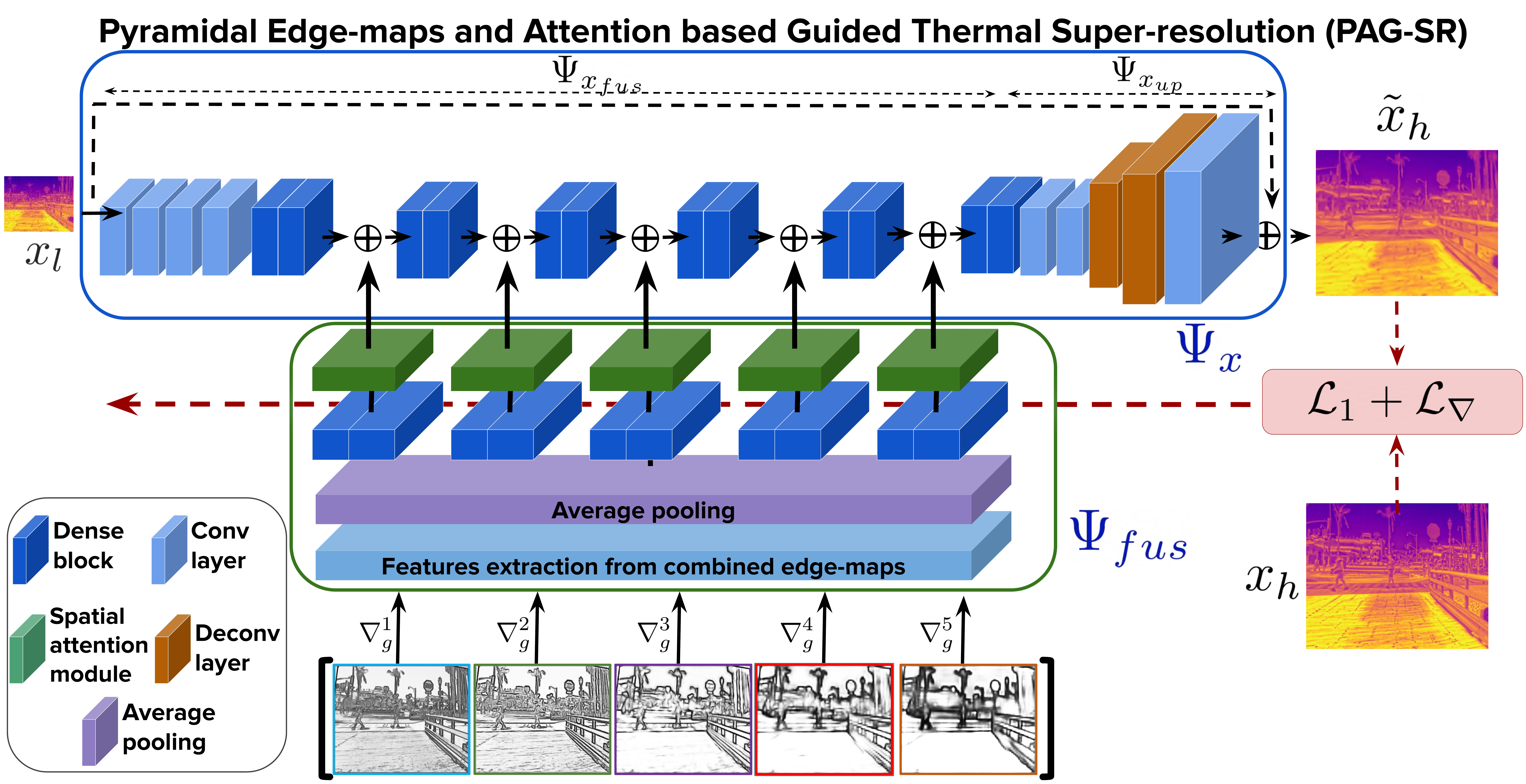}
\end{center}
\caption{Our method utilizes hierarchical edge information extracted from the guide visible image and integrates this information into our thermal super-resolution network at different network-depths with the help of feature extraction and self-attention mechanism. We progressively merge information obtained from the guide image and allow the network to adaptively deal with high-frequency information mismatch.}
\label{fig:mainarc}
\end{figure}

Figure \ref{fig:mainarc} shows the architecture of our proposed method. We denote the low-resolution thermal, the high-resolution visible and ground-truth thermal images as $\mathbf{x}_l$, $\mathbf{g}_h$ and $\mathbf{x}_h$, respectively. Our proposed network consists of two sub-networks: one for thermal image super-resolution, denoted as $\Psi_x$ and one for feature extraction and integration of multi-level edge-maps obtained from the visible image, denoted as $\Psi_{fus}$. Many existing edge-detection methods for single images~\cite{hed,rcf} extract multi-level edges and merge them to obtain object-level edge-map. They extract edges at different perceptual scales by taking output at different layers of the VGG~\cite{vgg} network. Consequently, to obtain edges having visible-range information at different perceptual scales, we used one of these existing methods~\cite{rcf}, which provides edge-maps at $5$ pyramidal levels. We denote these edge-maps as ${\nabla_g = [\nabla_g^1, \nabla_g^2, \dots, \nabla_g^5]}$.

To extract guidance information and fuse it into the super-resolution network, we propose a fusion network and denote it as $\Psi_{fus}$. As shown in Fig.~\ref{fig:mainarc}, $\Psi_{fus}$ first contains a convolution layer, which we denote as $\mathbf{C}_{edge}$. The convolution layer takes a concatenate of the multi-level edge-maps as an input and extracts features from these edge-maps collectively. We call these extracted features as edge-features and denote them as $\mathbf{G}_{edges} = \mathbf{C}_{edge} \circledast \mathbf{\nabla_g}$. $\mathbf{G}_{edges}$ contains the multi-level high-frequency guidance information from the visible-range image. These edge-features are then passed through an average pooling layer to reach the spatial resolution of $\mathbf{x}_l$. We perform the fusion at the low-resolution scale because our experiments showed that downsampling reduces the edge-mismatch between the input images and leads to better performance as compared to performing fusion in the spatially high-resolution feature-space. 

The next part of $\Psi_{fus}$ contains a set of dense\cite{dense} and spatial attention blocks\cite{spatial_att}, which we collectively call as the fusion sub-block. For $n$ edge-maps, we have $n$ sets of fusion sub-blocks, denoted as $\Psi_{fus}^{~n}$ that lead to $n$ connections into the thermal super-resolution network. Each fusion sub-block contains a dense-block with $2$ convolutional layers for extracting the relevant features from $\mathbf{G}_{edges}$ for that particular connection. Each dense-block is followed by a spatial attention block~\cite{spatial_att} that adaptively transforms the guidance information and outputs weighted features based on the spatial correlation of different channels in the features. Our spatial-attention block is similar to the one proposed in \cite{spatial_att} and its architecture is shown in Figure \ref{fig:att}. The mathematical description of the module can be found in the supplementary paper. We tried different variations of the fusion network $\Psi_{fus}$, details of which are mentioned in Section~\ref{sec:fus_ablation}.

\begin{figure}
\begin{center}
\includegraphics[width=0.6\linewidth]{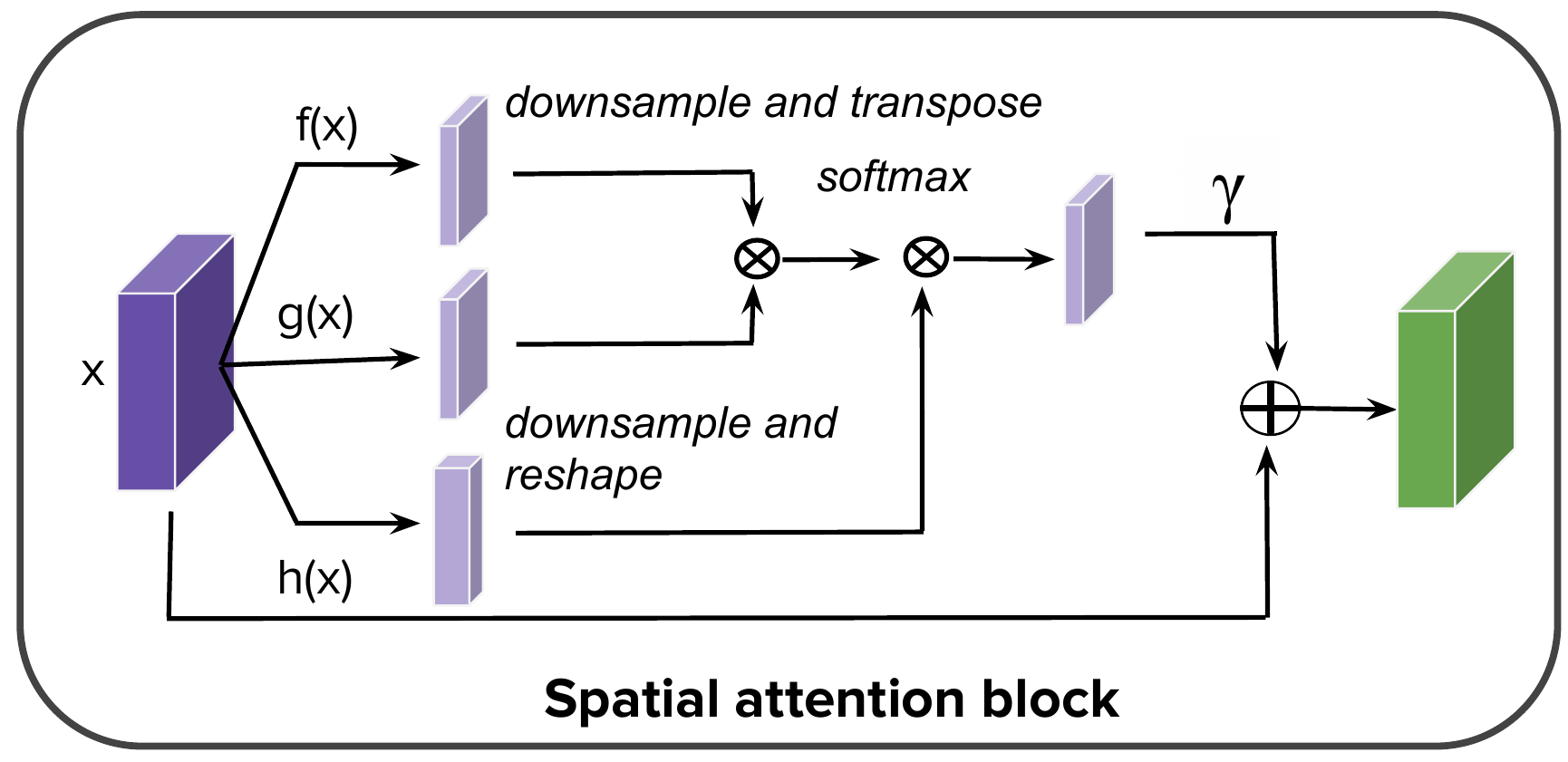}
\end{center}
  \caption{Architecture of the spatial attention block used in our fusion module. The module adaptively transforms the features according to spatial correlation inside each feature-map and outputs re-scaled features such that the relevant information has a higher activation.}
\label{fig:att}
\end{figure}

Our thermal image super-resolution sub-network, denoted as $\Psi_x$, consists of two parts: $\Psi_{x_{fus}}$ and $\Psi_{x_{up}}$, as shown in Fig. \ref{fig:mainarc}. The first part of the network, denoted as $\Psi_{x_{fus}}$, is the part that extracts information from the low-resolution thermal image and is merged with $\Psi_{fus}$ to receive the guidance information. $\Psi_{x_{fus}}$ contains convolutional layers having $32$ channels, which are followed by dense-blocks~\cite{dense} of two convolutional layers, each of which again have $32$ channels. For $n$ guide edge-maps, $\Psi_{x_{fus}}$ contains $n+1$ dense-blocks, denoted as $[\mathbf{D}_1, \mathbf{D}_2, \dots, \mathbf{D}_{n+1}]$.
The fusion operation can be summarised as:
\begin{equation}
    \mathbf{X}_{n+1} = \mathbf{D}_n(\mathbf{X}_n) + \Psi_{fus}^{~n}(\mathbf{G}_{edges})
\label{eq:fus}
\end{equation}
where, $\mathbf{X}_n$ and $\mathbf{D}_n$ are $n^{th}$ feature-map and dense-block of $\Psi_x$, respectively and $\Psi_{fus}^{~n}$ is the $n^{th}$ fusion sub-block, as mentioned in the previous paragraph. 

The features from $\Psi_{x_{fus}}$ are fed into $\Psi_{x_{up}}$, which contains convolutional and upsampling layers. For a ${2^k}$ super-resolution, $\Psi_{x_{up}}$ contains $k$ deconvolution layers. The output of $\Psi_{x_{up}}$ is $\mathbf{X}_{up} = \Psi_{{x_{up}}}(\mathbf{X}_{n+1})$. The final deconvolution layer is followed by a convolutional layer and a skip connection from the input image for residual learning. The output can be defined as:
\begin{equation}
    \mathbf{\tilde x}_h = \mathbf{C} \circledast \mathbf{X}_{up}   + \mathbf{x}_{l\uparrow} \equiv \Psi_x(\mathbf{x}_l)
\end{equation}

\noindent\textbf{Loss functions.} To learn the parameters of $\Psi_x$, our optimization function contains two loss terms. First is the reconstruction loss $\mathcal{L}_1(\mathbf{\tilde x}_h, \mathbf{x}_h) = ||\mathbf{\tilde x}_h - \mathbf{x}_h||_1$ for supervised training.
The second term is a gradient loss $\mathcal{L}_\nabla(\mathbf{\tilde x}_h, \mathbf{x}_h) = || \nabla (\mathbf{\tilde x}_h) - \nabla(\mathbf{x}_h)||_1 $ to explicitly penalize loss of high frequency details. Here $\nabla$ is the Laplacian operator that calculates both horizontal and vertical gradients. Hence, our overall loss function is
\begin{equation}
    \mathcal{L}(\mathbf{x}_l, \mathbf{x}_h, \Psi_{x}) 
     = \gamma_1 \mathcal{L}_1(\mathbf{\tilde x}_h, \mathbf{x}_h) + \gamma_2 \mathcal{L}_\nabla(\mathbf{\tilde x}_h, \mathbf{x}_h)
\end{equation}
\noindent We found the optimal values for $\gamma_1$ and $ \gamma_2 $ to be $10$ and $1$, respectively.

\section{Experiments}
\subsection{Datasets and Setup}

We perform experiments on three datasets: FLIR-ADAS~\cite{flir}, CATS~\cite{cats} and KAIST~\cite{kaist}. FLIR-ADAS contains unrectified stereo thermal and visible-range image pairs having a resolution of $512\times640$ and $1600\times1800$, respectively. Since the dataset does not contain any calibration images, we rectified one image pair manually, by identifying the correspondences and estimating the relative transformation matrices. We used this estimated transformations to rectify rest of the images in the dataset. After rectification, both thermal and visible-range images are of resolution $512\times640$. The CATS dataset contains rectified thermal and visible images, both of dimensions $480\times640$. This dataset also contains ground-truth disparity maps between the two images, but we observed that the disparity-maps are not accurate and results in artifacts when either of the images is warped using the disparity. Therefore, we used the rectified yet unaligned image pairs for our experiments, similar to the FLIR-ADAS dataset.  Hence, both datasets have rectified image pairs (i.e. epipolar lines are horizontal), but they are not pixel-wise aligned. In contrast, the third KAIST dataset contains aligned thermal and visible images of resolution $512\times640$. This dataset was captured using a beam-splitter based setup and hence it has less practical similarity with the low-resolution thermal cameras. We therefore use this dataset for a smaller set of GSR methods, to present a baseline comparison on aligned thermal and visible images.


To create the low-resolution dataset, we used the blur-downscale degradation model proposed in \cite{blur-sr} to simulate the low-resolution images. For the training-set, we down-sample images using blur kernels with ${\sigma}\in[0,4]$ at a step of $0.5$. For the FLIR-ADAS dataset, our training set contains 43830 image pairs and the test-set contains 1257 pairs. In the CATS dataset, our training set contains 944 image-pairs and the test-set contains 50 pairs. Since CATS training-set is quite small, we used to it fine-tune the models pre-trained on FLIR-ADAS training-set and then tested them on the CATS test-set. For the KAIST dataset, our training-set contains $5581$ image-pairs and our test-set contains $964$ pairs. We perform experiments for $\times4$ and $\times8$ upsampling factors. For both datasets, the input thermal-resolution is close to the resolution of low-cost thermal cameras like FLIR AX8.  The $\mathbf{x}_l$ dimensions are ${64\times80}$ for FLIR-ADAS and KAIST; and ${60\times80}$ for CATS. Hence, for $\times4$ SR, the guide image and the ground-truth image for FLIR-ADAS and KAIST are of resolution ${256\times320}$ and for CATS, they are of resolution ${240\times320}$. Similarly, for $\times8$, the corresponding guide and GT images are of sizes ${512\times640}$ and ${480\times640}$. For network optimization, we use ADAM optimizer \cite{adam} with a learning rate of $1\times10^{-4}$. The experiments were performed on a Nvidia 2080Ti GPU.

\begin{table}[t]
\centering
\resizebox{0.87\linewidth}{!}{
\begin{tabular}{l|c|c|cccc}
\toprule
\textbf{Method} & \hspace{0.3cm} \textbf{G/S} \hspace{0.3cm} & \textbf{Scale} & \textbf{PSNR} & \textbf{SSIM} & \textbf{MSE} & \textbf{LPIPS}\\
\hline
\hline
Bicubic & Single &  $\times4$ & 28.37 & 0.890 & 0.001652 & 0.405\\
RDN \cite{rdn} & Single & $\times4$ & 29.28 &  0.906 &  0.001441 & 0.282 \\
RCAN \cite{rcan} & Single & $\times4$ & 29.18 & \underline{0.908} & 0.001483 & 0.228 \\
SAN \cite{san} & Single & $\times4$ & 26.47 & 0.859 & 0.002567 & 0.229 \\
TGV2-L2 \cite{aniso} & Guided & $\times4$ & 28.77 & 0.892 & 0.001601 & 0.422 \\
FBS \cite{fbs} &  Guided & $\times4$ & 25.48 & 0.787 & 0.003152 & 0.387 \\
Joint-BU \cite{joint} &  Guided & $\times4$ & 27.77 & 0.874 & 0.001855 & 0.284\\
Infrared SR \cite{ir} &  Guided & $\times4$ & 28.21 & 0.889 & 0.001692 & 0.405  \\
SDF \cite{sdf} &  Guided & $\times4$ & 28.70 & 0.875 & 0.001488 & 0.321 \\
MSF-SR \cite{msf} &  Guided & $\times4$ & 29.21 &  0.901 & 0.001447 & 0.200 \\
MSG-Net \cite{msgnet} &  Guided & $\times4$ & \underline{29.46} & 0.897 & \underline{0.001341} & \underline{0.184}\\ 
PixTransform \cite{pixt} &  Guided & $\times4$ & 24.84 & 0.787 & 0.003679 & 
0.329 \\
Deep-ISTA \cite{ista} &  Guided & $\times4$ & 25.86 & 0.828 & 0.028939 & 0.529 \\
PAG-SR (Ours) & Guided & $\times4$ & \textbf{29.56} & \textbf{0.912} & \textbf{0.001309} & \textbf{0.147} \\
\bottomrule
\toprule
RDN \cite{rdn} & Single & $\times8$ & 26.80 & 0.833 & 0.002314 & 0.389\\
RCAN \cite{rcan} & Single & $\times8$ & 22.35 & 0.758 & 0.006771 & 0.414\\
SAN \cite{san} & Single & $\times8$ & 25.38 & 0.811 & 0.003251 & 0.536 \\
TGV2-L2 \cite{aniso} & Guided & $\times8$ & 26.42 & 0.821 & 0.002526 & 0.399  \\
FBS \cite{fbs} &  Guided & $\times8$ & 25.03 & 0.770 & 0.003451 & 0.476 \\
Joint-BU \cite{joint} &  Guided & $\times8$ & 25.61 & 0.803 & 0.003006 & 0.406 \\
Infrared SR \cite{ir} &  Guided & $\times8$ & 26.03 & 0.817 & 0.002782 & 0.521 \\
SDF \cite{sdf} &  Guided & $\times8$ & 26.72 & 0.819 & 0.002379 & 0.363 \\
MSF-SR \cite{msf} &  Guided & $\times8$ & \underline{27.92} & 0.835 & 0.002350 & \underline{0.249} \\
MSG-Net \cite{msgnet} &  Guided & $\times8$ & 27.29 & 0.827 & \underline{0.002263} &  0.296 \\ 
PixTransform \cite{pixt} &  Guided & $\times8$ & 23.31 & \underline{0.836} & 0.005224 & 0.371 \\
Deep-ISTA \cite{ista} &  Guided & $\times8$ & 25.56 & 0.778 & 0.030982 & 0.598 \\
PAG-SR (Ours) & Guided & $\times8$ & \textbf{28.77} & \textbf{0.919} & \textbf{0.001581} & \textbf{0.214} \\
\hline
\bottomrule
\end{tabular}}
\caption{Comparison of existing methods on FLIR-ADAS for $\times4$ and $\times8$ SR cases.} 
\label{table:flir}
\end{table}

\subsubsection{Comparison details.} 
We compare our method with $9$ existing GSR methods: TGV2-L2 \cite{aniso}, FBS \cite{fbs}, Joint-BU \cite{joint}, Infrared-SR \cite{infra}, SDF \cite{sdf}, MSF-STI-SR \cite{msf}, MSG-Net \cite{msgnet}, Pix-Transform \cite{pixt} and Deep-ISTA \cite{ista}. We also include comparison with a few recent single image SR methods such as RCAN \cite{rcan}, RCAN \cite{rcan} and SAN \cite{san}. We used the publicly available codes for the existing methods and trained the CNN based single and guided SR methods on the corresponding thermal datasets to perform the comparison. For the CATS dataset, the models pre-trained on FLIR-ADAS dataset were used for fine-tuning. We kept the default settings for most of the methods, except for few filtering based methods such as FBS and Joint-BU, where the weights had to be adjusted to reconstruct better texture-details in the super-resolved images. 

\subsubsection{Metrics.} We use four metrics to quantitatively assess the reconstructions: PSNR, SSIM, Mean-squared Error (MSE) and Perceptual distance (LPIPS) \cite{perceptual}. Among these, PSNR and SSIM are distortion-based metrics and hence, can be biased towards smooth or blurred images. Therefore, we also use LPIPS, a perceptual metric that computes the perceptual distance between the reconstructed and the ground-truth images. A point to note is that since the reconstructed images are thermal measurements, better MSE is also an important factor while comparing the methods.

\begin{table}[!htbp]
\centering
\resizebox{0.87\linewidth}{!}{
\begin{tabular}{l|c|c|cccc}
\toprule
\textbf{Method} & \hspace{0.3cm} \textbf{G/S} \hspace{0.3cm} & \textbf{Scale} & \textbf{PSNR} & \textbf{SSIM} & \textbf{MSE} & \textbf{LPIPS}\\
\hline
\hline
Bicubic & Single & $\times4$ & 32.19 & 0.959 & 0.000744 & 0.395\\
RDN \cite{rdn} & Single & $\times4$ & 29.41 & 0.914 & 0.004811 & 0.357 \\
RCAN \cite{rcan} & Single &  $\times4$ & 31.89 & \underline{0.966} & 0.000796 &  0.159 \\
SAN \cite{san} & Single & $\times4$ & 33.41 & 0.960 & 0.000507 & \textbf{0.141} \\
TGV2-L2 \cite{aniso} & Guided & $\times4$ &  32.17 & 0.938 & 
0.000741 & 0.225 \\
FBS \cite{fbs} & Guided & $\times4$ & 29.12 & 0.825 & 0.035104 & 0.450 \\
Joint-BU \cite{joint} & Guided & $\times4$ & 31.23 & 0.953 & 0.000914 & 0.233 \\
Infrared SR \cite{ir} & Guided & $\times4$ & 28.27 & 0.901 & 0.031501 &  0.348 \\
SDF \cite{sdf} & Guided & $\times4$ & 32.56 & 0.941 & 0.000686 & 0.246 \\
MSF-SR \cite{msf} & Guided & $\times4$ & 29.37 & 0.830 & 0.022598 & 0.415 \\
MSG-Net \cite{msgnet} & Guided & $\times4$ & 31.56 & 0.964 & 0.000789&0.177 \\ 
PixTransform \cite{pixt} & Guided & $\times4$  & 28.48 & 0.792 & 0.185427& 0.442 \\
Deep-ISTA \cite{ista} & Guided & $\times4$ & \underline{33.72} & 0.956 & \underline{0.000488} & 0.178 \\
PAG-SR (Ours) & Guided & $\times4$ & \textbf{34.97} & \textbf{0.968} & \textbf{0.000461} & \underline{0.161} \\
\bottomrule
\toprule
Bicubic & Single & $\times8$ & 31.45 &  0.958 & 0.000868 & 0.413\\
RDN \cite{rdn} & Single & $\times8$ & \underline{33.31} & 0.956 & 0.000631 & 0.392\\
RCAN \cite{rcan} & Single & $\times8$ & 27.63 & 0.931 & 0.002296 & 0.332\\
SAN \cite{san} & Single & $\times8$ & 32.17 & \underline{0.953} & 0.000615 & 0.278\\
TGV2-L2 \cite{aniso} & Guided & $\times8$ & 31.55 & 0.951 &0.000846 &
0.303 \\
FBS \cite{fbs} & Guided & $\times8$ & 29.03 & 0.855 & 0.035654 & 0.495 \\
Joint-BU \cite{joint} & Guided & $\times8$ & 30.21 & 0.950 & 0.001131 & 0.314 \\
Infrared SR \cite{ir} & Guided & $\times8$ & 25.23 & 0.904 & 0.029725& 0.409\\
SDF \cite{sdf} & Guided & $\times8$ & 31.91 & 0.948 & 0.000778 & 0.319 \\
MSF-SR \cite{msf} & Guided & $\times8$ &  27.97 & 0.811 & 0.032487 & 0.418 \\
MSG-Net \cite{msgnet} & Guided & $\times8$ & 32.83 & 0.957 & 0.000622 & \underline{0.270}\\ 
PixTransform \cite{pixt} & Guided & $\times8$ & 27.79 & 0.783 & 0.121195& 0.584\\
Deep-ISTA \cite{ista} & Guided & $\times8$ & 32.51 &0.949 & \underline{0.000595} & 0.290 \\
PAG-SR (Ours) & Guided & $\times8$ & \textbf{33.18} & \textbf{0.963} & \textbf{0.000537} & \textbf{0.246}\\
\hline
\bottomrule
\end{tabular}}

\vspace{0.3cm}
\caption{Comparison of guided super-resolution methods on CATS dataset for $\times4$ and $\times8$ upsampling cases. \textit{Higher PSNR, SSIM and lower MSE, LPIPS are better.}} 
\label{table:cats}
\end{table}

\subsection{Quantitative comparison} 

Table~\ref{table:flir} and \ref{table:cats} show the results for $\times4$ and $\times8$ upsampling factors for FLIR-ADAS and CATS datasets, and Table \ref{table:kaist} shows the results for $\times4$ SR on KAIST dataset. A general trend among the existing methods is that they perform quite well in terms of distortion metrics but poorly in terms of the perceptual metric for the FLIR and CATS datasets. For these datasets, MSG-Net~\cite{msgnet} and MSF-SR~\cite{msf} results are the closest to our method. PixTransform~\cite{pixt} and Deep-ISTA~\cite{ista} are the most recent methods, yet they perform poorly as compared to the others for the FLIR and CATS datasets, mostly due to edge-mismatch and inability to accommodate the texture difference. For the FLIR-dataset, we observe a small variance in the metric values but for the CATS dataset, the variance is much higher. The reason for this is the higher disparity range or higher misalignment between the input images from the CATS dataset. In contrast, the KAIST dataset has aligned images and hence, the results show much less variance as compared to the other two datasets. For this dataset, Deep-ISTA seems to be the closest to our method in terms of performance.

Our method outperforms the existing methods in terms of distortion as well as perceptual metrics on all three datasets. We observe a significant margin between ours and the existing methods' performances, especially in the case of $\times8$ SR on FLIR-ADAS dataset. In Table \ref{table:cats}, we observe a similar pattern in the metric values. However, an interesting observation is that in the $\times4$ case for CATS dataset, the single image SR methods perform better than many GSR methods. We believe this could be due to the edge-mismatch caused by the misalignment.  For the KAIST dataset, our results are better than the existing methods, which indicates that our method works better for both aligned and misaligned inputs. \\

\begin{table}[!htbp]
\centering
\begin{tabular}{l|c|c|cccc}
\toprule
\textbf{Method} & \hspace{0.3cm} \textbf{G/S} \hspace{0.3cm} & \textbf{Scale} & \textbf{PSNR} & \textbf{SSIM} & \textbf{MSE} & \textbf{LPIPS}\\
\hline
SDF \cite{sdf} & Guided & $\times4$ & 24.14 & 0.830 & 0.005647 & 0.242 \\
MSF-SR \cite{msf} & Guided & $\times4$ & 25.52 & 0.855 & 0.004319 & 0.274 \\
MSG-Net \cite{msgnet} & Guided & $\times4$ & 25.61 & {0.868} & 0.004031 & 0.242 \\ 
PixTransform \cite{pixt} & Guided & $\times4$ & 24.85 & 0.828 & 0.007692 & 0.325 \\
Deep-ISTA \cite{ista} & Guided & $\times4$ & 26.12 & \textbf{0.871} & 0.003846 & 0.273 \\
PAG-SR (Ours) & Guided & $\times4$ & \textbf{27.98} & 0.856 & \textbf{0.002399} & \textbf{0.219} \\
\bottomrule
\end{tabular}
\vspace{0.3cm}
\caption{Comparison of GSR methods on KAIST dataset for $\times4$ upsampling case.} 
\label{table:kaist}
\end{table}

\begin{figure}
\begin{center}
\footnotesize{(a) Visual comparison for $\times4$ guided super-resolution}
\includegraphics[width=\linewidth]{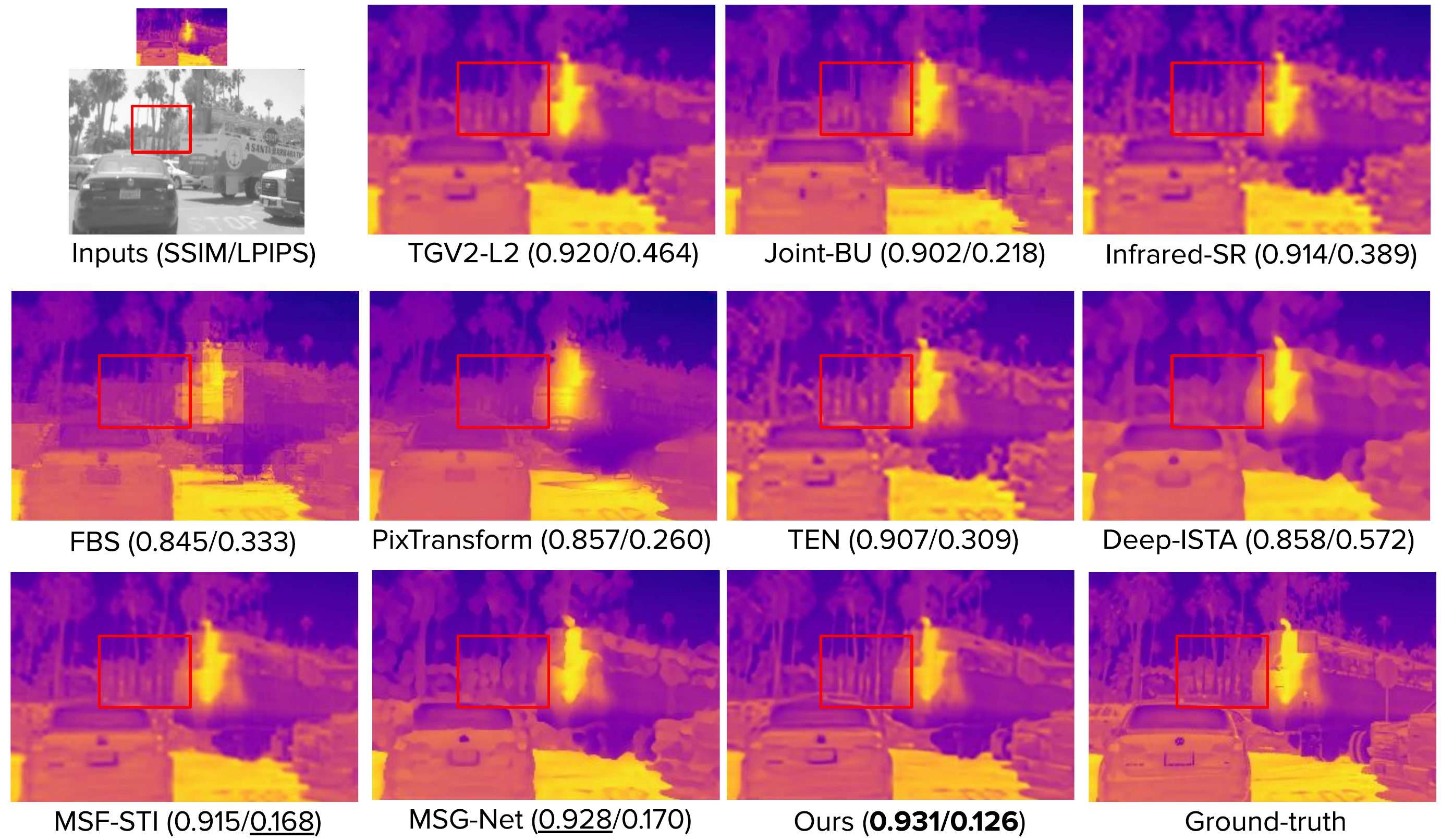}\\
\vspace{0.3cm}
\footnotesize{(b) Visual comparison for $\times8$ guided super-resolution}
\includegraphics[width=\linewidth]{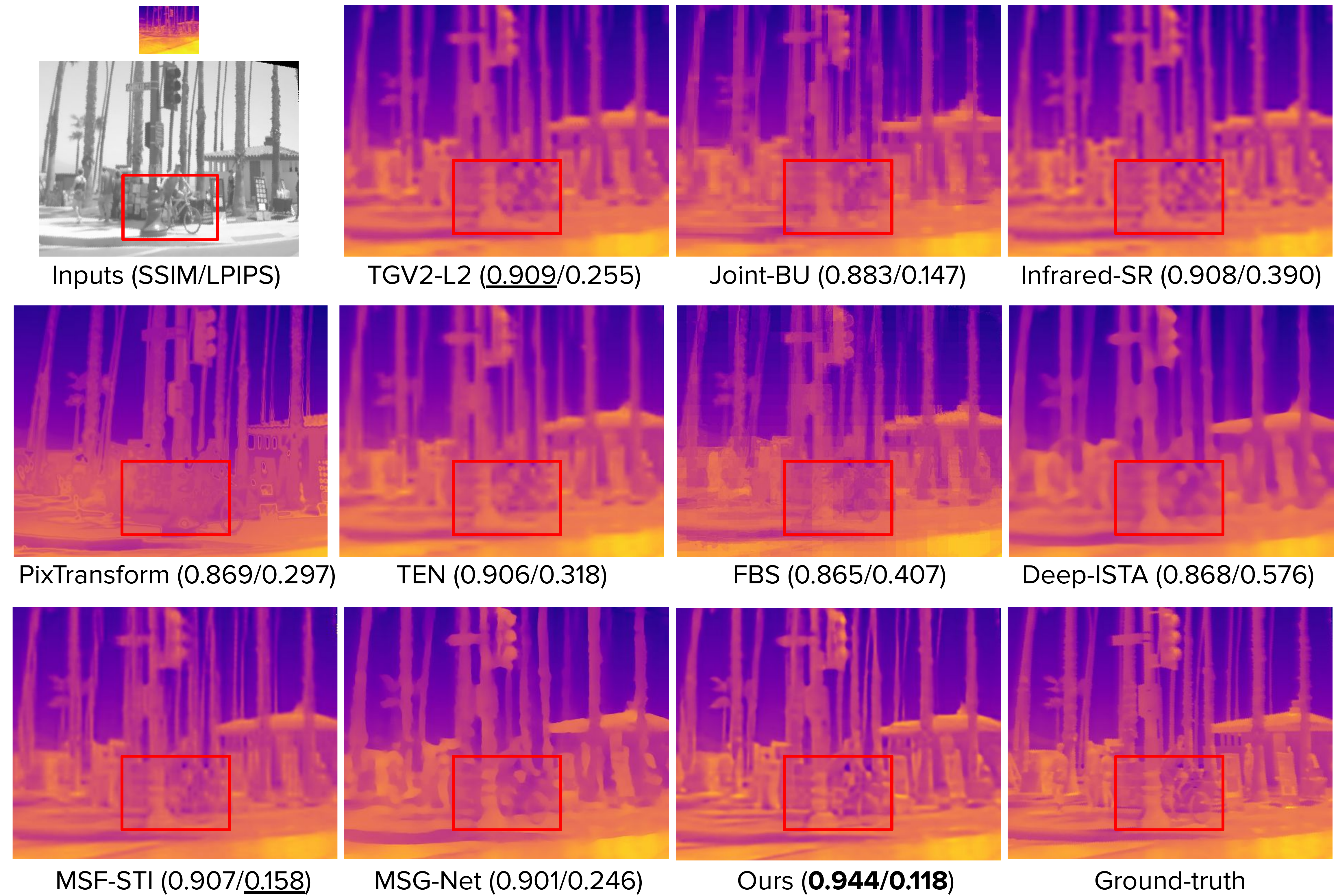}
\end{center}
  \caption{Visual comparison on sample images from FLIR-ADAS dataset. Our method reconstructs high frequency details more accurately and has less artifacts due to mismatched edges as compared to existing GSR methods, for \textit{e.g.} the ghosting effect of trees in MSF-STI's $\times8$ reconstruction or blurred edges in most of the other methods. We also achieve higher SSIM and lower perceptual distance (LPIPS) values. }
\label{fig:x8}
\end{figure}

\begin{figure}[t]
\centering
\includegraphics[width=\linewidth]{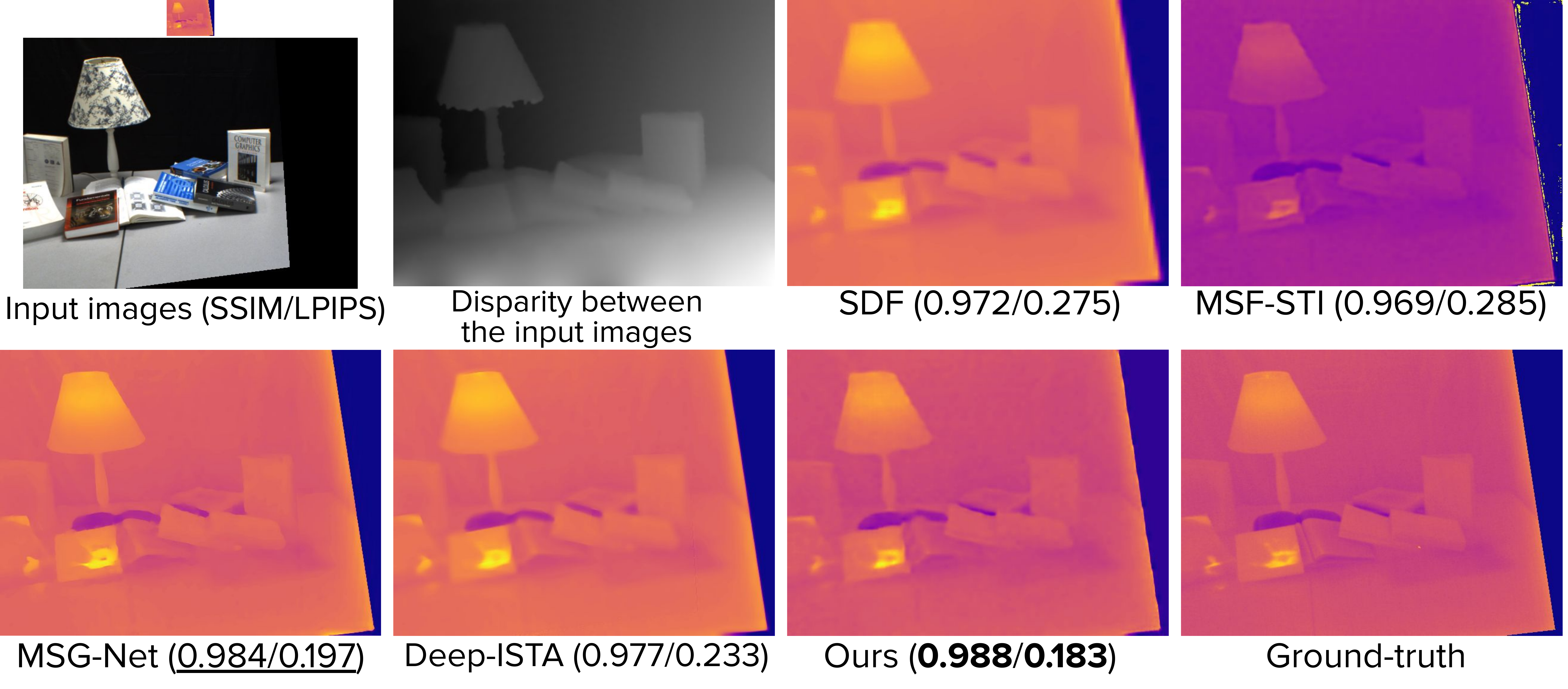}\\
\vspace{0.1cm}
\includegraphics[width=\linewidth, trim = 0 0.5cm 0 0, clip]{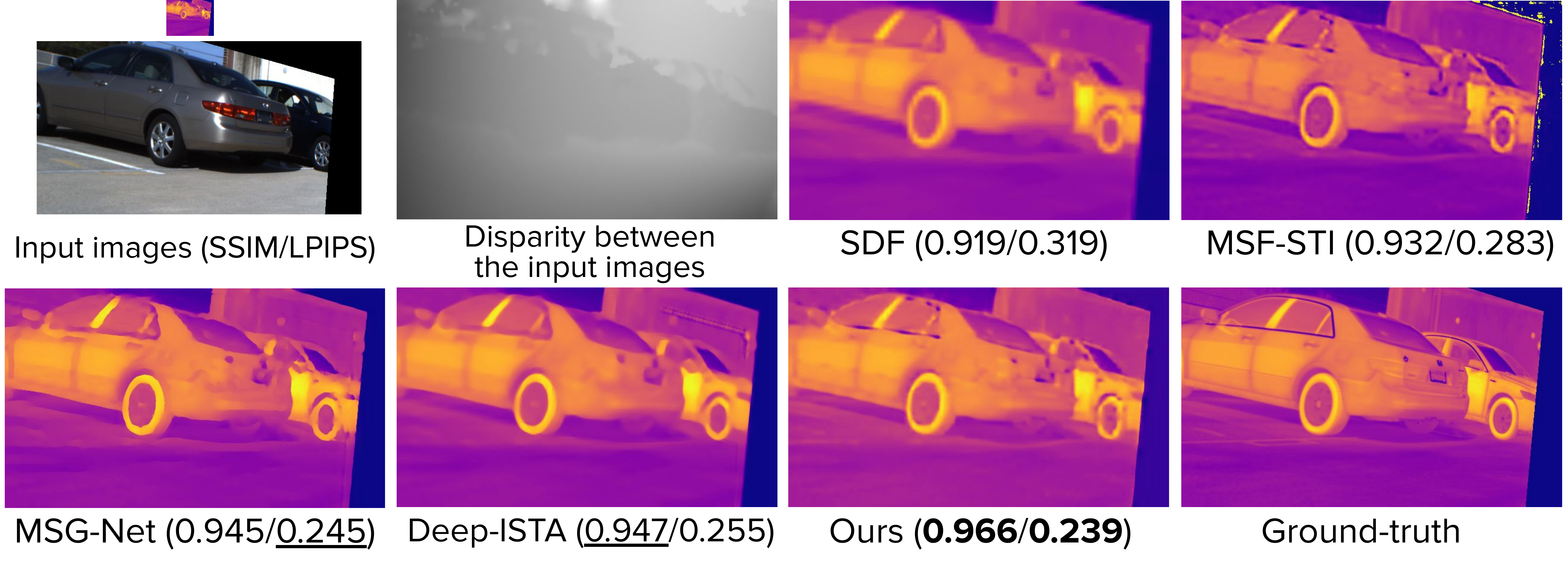}
\vspace{-0.6cm}
  \caption{
  Visual comparison on images from CATS dataset for $\times8$ SR. 
}
\vspace{-0.5cm}
\label{fig:cats}
\end{figure}

\noindent \textbf{Robustness towards texture-mismatch and misalignment.} The CATS dataset has higher misalignment which is visualized in the disparity-maps in Fig.\ref{fig:cats}. Misalignment results in higher texture-mismatch, which consequently reduces the performance of many existing methods. We speculate that usage of edge-maps and attention module in our method provides some form of robustness towards misalignment and hence is the cause of our better performance. 
To validate this, we experimented with a variant of our network that takes an RGB guide input and does not have the attention block inside the fusion sub-blocks. We found that this model performs lower 
than our proposed method which has pyramidal edge-maps and attention based fusion. In terms of metrics, the model without edge-maps and attention achieved an SSIM of 0.901 and LPIPS of 0.173 as compared our proposed model's SSIM of 0.912 and LPIPS of 0.147, for $\times4$ SR on FLIR dataset. This indicates that the edge-maps and attention module contribute towards the performance improvement and hence, are able to tackle texture-mismatch and most probably misalignment as well to a certain extent. 

\subsection{Qualitative comparison}
We show the qualitative comparison of our method for $\times4$ and $\times8$ upsampling rates on FLIR-ADAS dataset in Fig.~\ref{fig:x8}. Most of the existing methods have blurred edges, especially in the case of $\times8$ SR, which could be either due to very low-resolution of the input thermal images ($60\times80$) or due to texture-mismatch and improper propagation of guidance information. Among the existing methods, MSF-STI, MSG-Net and Joint-BU results are considerably good, yet our method reconstructs high-frequency details more faithfully and has much sharper edges. For \textit{e.g.} in Fig. 4(a), the branches in the red inset for our result are evidently most clear as compared to the other methods. Moreover, ghosting artifacts can be found in the results from few existing methods, such as MSF-STI's reconstruction for $\times8$ case in Fig 4(b). In contrast, our method is able to reconstruct the object structure better for the same image and does not have such artifacts. 

Fig. \ref{fig:cats} shows the $\times8$ super-resolution results for few samples from the CATS dataset, that has higher misalignment as compared to the FLIR-ADAS dataset. We can observe that the existing guided super-resolution methods show more blur as compared to FLIR-ADAS dataset, mostly due to increased texture-mismatch caused by a higher misalignment. However, our results are overall much sharper than the existing methods, which indicates a comparatively higher robustness towards misalignment, as mentioned in Section 4.2. 

\subsection{Ablation studies}
\label{sec:ablation}

\subsubsection{Usefulness of edge-maps.} 
Ideally, the RGB image could be used to extract features/edges in an end-to-end manner. However, estimating the optimal edge-map is very challenging because the SR network will require object-level awareness while extracting features. Using multilevel edge-maps simplifies this task by providing the object-edge information explicitly. To validate this hypothesis, we performed some experiments where we replaced the edge-maps with the visible RGB image for few variants of our model. The experiment was performed for $\times8$ SR on FLIR-ADAS dataset. Our proposed PAG-SR contains guide information fusion at $5$ positions to the thermal super-resolution network. However, since edge-features $\mathbf{G}_{edges}$ contain information from all the edge-maps, these positions can be reduced or expanded. Table~\ref{table:rgb_comparison} summarizes the results for a couple of such variations. 
When the guide information is fed at position 1, namely after first dense-block or $\mathbf{X}_1$, then RGB performs slightly better than the edge-maps. However, in the case of fusion at $5$ positions, the edge-maps perform better than both RGB, RGB combined with edge-maps, and all other variants as well. 
Thus, we can conclude that using edge-maps as a guidance information and fusing them at multiple-positions is overall a better strategy. We also performed a study to analyze the contribution of different levels of the edge-maps towards performance. The results can be found in Table 1 of the supplementary paper.

\begin{table}[t]
\centering
\begin{tabular}{c@{\extracolsep\fill}c@{\extracolsep\fill}c@{\extracolsep\fill}c@{\extracolsep\fill}c@{\extracolsep\fill}|cc|ccc}
\toprule
\multicolumn{5}{c|}{\textbf{Integration positions}} & \multicolumn{2}{c|}{\textbf{Input type}} & \multirow{2}{*}{PSNR} & \multirow{2}{*}{SSIM} & \multirow{2}{*}{LPIPS} \\
\hspace{0.3cm} 1 \hspace{0.3cm} & 2 \hspace{0.3cm} & 3 \hspace{0.3cm} & 4 \hspace{0.3cm} & 5 \hspace{0.3cm} & RGB & Edge-maps & \\
\hline
\checkmark &&&&& \checkmark & & 28.05 & 0.916 & 0.249\\
\checkmark &&&&& & \checkmark & 28.34 & 0.909 & 0.222 \\
\checkmark & \checkmark & \checkmark &&& \checkmark &&  28.19 & 0.916 &  0.243 \\
\checkmark & \checkmark & \checkmark &&&& \checkmark & \textbf{28.85} & 0.916 & 0.215 \\
\checkmark & \checkmark & \checkmark & \checkmark & \checkmark& \checkmark &  & 28.14 &  0.916 & 0.258 \\
\checkmark & \checkmark & \checkmark & \checkmark & \checkmark && \checkmark & 28.77 & \textbf{0.919} & \textbf{0.214}  \\
\checkmark & \checkmark & \checkmark & \checkmark & \checkmark & \checkmark & \checkmark &  28.83 &  0.915 & 0.221 \\
\hline
\bottomrule 
\end{tabular}
\caption{Performance variation with respect to guidance input type: RGB \textit{vs} edge-maps for $\times8$ super-resolution on FLIR-ADAS dataset.}
\label{table:rgb_comparison}
\end{table}

\begin{table}[!htbp]
\centering
\begin{tabular}{c|c|c|ccc}
\toprule
\textbf{Edge-features($G_{edges}$)} & \textbf{Dense block} & \textbf{Attention module} &  {PSNR} & {SSIM} & LPIPS \\
\hline
\xmark & \xmark & \xmark & {27.95} & {0.837} & \textbf{0.213}\\
\checkmark & \xmark & \xmark &  28.15 & 0.904 & 0.223\\
\checkmark & \xmark & \checkmark & 28.39 &  \underline{0.910} & \underline{0.214}\\
\checkmark & \checkmark & \xmark & \textbf{28.87} & 0.907 & 0.221 \\
\checkmark & \checkmark & \checkmark & \underline{28.77} & \textbf{0.919} & \underline{0.214} \\
\hline
\bottomrule 
\end{tabular}
\caption{Performance of different variations of our guidance fusion module.}
\vspace{-0.5cm}
\label{table:model_comparison}
\end{table}

\subsubsection{Contribution of different components of the fusion network}
\label{sec:fus_ablation}
\begin{figure}[!htbp]
\centering
\includegraphics[width=0.9\linewidth]{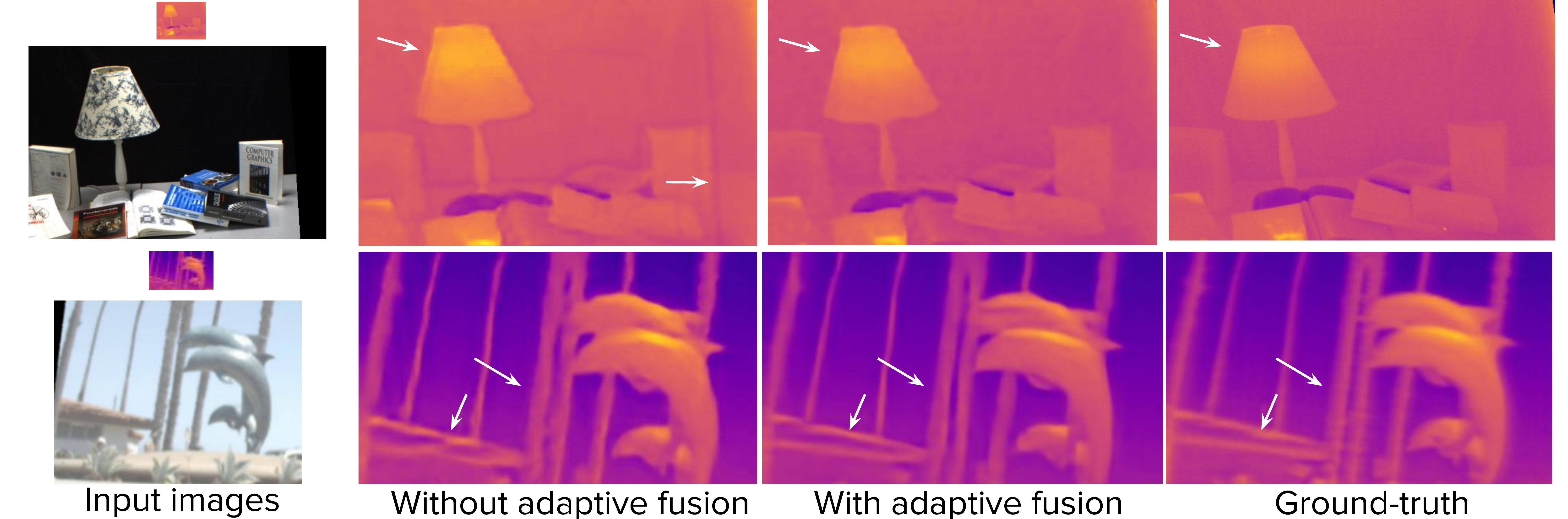}
\vspace{-0.2cm}
  \caption{Comparison of results from our method with and without the fusion module.}
\vspace{-0.5cm}
\label{fig:simple_vs_adv}
\end{figure}

To analyze the contribution of different components of our fusion network, we computed the performance of some variants of our fusion network while keeping the SR network constant for $\times8$ SR on FLIR-ADAS dataset. The results of the experiments are summarized in Table~\ref{table:model_comparison}. The simplest model is our proposed method without the fusion module and hence contains edge-maps directly added to the SR network. Other variants include either having the dense-blocks or the attention module or neither or both in each fusion sub-block. The results show that having both dense-blocks and attention module helps is achieving better reconstructions. Moreover, Fig.~\ref{fig:simple_vs_adv} shows a visual comparison of the results from model with and without the fusion block. The model without the fusion module contains many artifacts induced by the edge-mismatch between the input images. Most of such artifacts are eliminated by our fusion network with the help of appropriate selection of edge-information.

\section{Conclusion}
We proposed a hierarchical edge-maps based guided super-resolution algorithm  that tackles edge-mismatch due to spectral-difference between the input low-resolution thermal and high-resolution visible-range images in a systematic and holistic manner. Our method robustly combines multi-level edge information extracted from the visible-range image into our tailored thermal super-resolution network with the help of attention based guidance propagation and consequently produces better high-frequency details. We showed that our results are significantly better both perceptually and quantitatively than the existing state-of-the-art guided super-resolution methods.



%
%
\bibliographystyle{splncs04}
\bibliography{egbib}
\end{document}